\documentclass{article}

\usepackage{PRIMEarxiv}

\usepackage[utf8]{inputenc} 
\usepackage[T1]{fontenc}    
\usepackage{hyperref}       
\usepackage{url}            
\usepackage{booktabs}       
\usepackage{amsfonts}       
\usepackage{nicefrac}       
\usepackage{microtype}      
\usepackage{lipsum}
\usepackage{fancyhdr}       
\usepackage{graphicx}       
\graphicspath{{media/}}     
\usepackage{tabularx}
\usepackage{booktabs}
\usepackage{float}

\title{Developing and Evaluating Tiny to Medium-Sized Turkish BERT Models}

\author{
  Himmet Toprak Kesgin, Muzaffer Kaan Yüce, Mehmet Fatih Amasyali \\
  \textit{Cosmos Research Lab.} \\
  \textit{Department of Computer Engineering} \\
  \textit{Yildiz Technical University} \\
  Istanbul, Turkey\\
  \texttt{\{tkesgin, kaan.yuce, amasyali\}@yildiz.edu.tr}
}

\begin{document}
\maketitle

\begin{abstract}
This study introduces and evaluates tiny, mini, small, and medium-sized uncased Turkish BERT models, aiming to bridge the research gap in less-resourced languages. We trained these models on a diverse dataset encompassing over 75GB of text from multiple sources and tested them on several tasks, including mask prediction, sentiment analysis, news classification, and, zero-shot classification. Despite their smaller size, our models exhibited robust performance, including zero-shot task, while ensuring computational efficiency and faster execution times. Our findings provide valuable insights into the development and application of smaller language models, especially in the context of the Turkish language.
\end{abstract}


\section{Introduction}
As natural language processing (NLP) continues to evolve, the application and fine-tuning of language models such as BERT (Bidirectional Encoder Representations from Transformers) \cite{devlin2018bert} have demonstrated promising advancements in various linguistic tasks. Developed by researchers at Google, BERT's bidirectional training of transformers has shown remarkable success in understanding the intricacies of language semantics. However, the application of BERT models in less-resourced languages such as Turkish still remains a somewhat uncharted territory.

This gap in research becomes particularly pronounced when examining small-sized versions of BERT models. While large and base BERT models have been extensively studied and implemented, their smaller counterparts offer unique value in terms of computational efficiency and ease of deployment, particularly for languages with fewer linguistic resources available. However, to the best of our knowledge, there is currently no published research on the effectiveness of smaller Turkish BERT models, a gap that this study aims to fill.

In this paper, we present for the first time, the creation and evaluation of tiny, mini, small, and medium-sized Turkish BERT models, all uncased. Our models were trained on a substantial dataset, aggregating over 75GB of text from various sources, including Turkish version of Oscar \cite{abadji2022towards} and MC4 \cite{raffel2020exploring} datasets, diverse news datasets, novels, and the latest data from Wikipedia \cite{musabgwikipediatr}. This breadth and diversity in training data underpin our models' robust performance across several tasks.

We have tested these models on a variety of tasks, including mask prediction, sentiment analysis, news classification and zero shot classification, and present their respective accuracy scores and execution times. Notably, our results show that these smaller models provide much faster prediction times due to their reduced computational requirements, making them particularly practical for real-world applications.

The intention behind minimizing the size of BERT models is to make the state-of-the-art language model more accessible for tasks necessitating quicker processing and less memory, with minimal loss of performance. Our results offer valuable insights for both researchers and practitioners in the NLP field, particularly within the Turkish language context. This paper also paves the way for future development of smaller, more efficient language models.

In an effort to contribute to the broader scientific and AI communities, all models utilized and fine-tuned in this research will be shared publicly on the Hugging Face Model Hub. This platform is a large-scale repository for pre-trained models, which facilitates sharing and collaboration among researchers and developers in the field of NLP. By sharing our models, we aim to promote transparency in our research process and provide a resource for others to replicate our experiments, scrutinize our methodology, and build upon our findings.

\section{Literature Review}
The extensive application of BERT models for Turkish NLP tasks, as exemplified by BERTurk \cite{stefan_schweter_2020_3770924}, underscores the effectiveness of large-scale transformer models in language-specific tasks. However, the computational intensity of these models necessitates efficient alternatives, like DistilBERTurk \cite{distilberturk}, which leverage knowledge distillation techniques for superior performance with reduced computational demands. In our experiments, we used the BERTurk model as the baseline, referred to as the 'Base' model.

Research aimed at making BERT models smaller and faster has been quite successful. Sanh et al. (2019) proposed DistilBERT \cite{sanh2019distilbert}, a smaller version of BERT that retained 97\% of the language understanding capabilities of its larger counterpart. They achieved this reduction by leveraging knowledge distillation during the pre-training phase.

Echoing these findings, Jiao et al. (2020) presented TinyBERT \cite{jiao2019tinybert}, which employs a two-stage learning framework for distillation during both pre-training and task-specific learning stages. Despite being significantly smaller, TinyBERT demonstrated comparable performance to its larger counterparts on the GLUE benchmark.

Similarly, Tsai et al. (2020) proposed a compact model suitable for sequence labeling tasks that was 6x smaller and 27x faster than the standard multilingual BERT \cite{tsai2019small}. The model exceeded the performance of state-of-the-art multilingual baselines, showing particular strength with low-resource languages.

In contrast to the predominantly English-centered research, Ulcar et al. (2019) offer an in-depth analysis of precomputed ELMo embeddings for seven languages, emphasizing the importance of the size of training sets and the superiority of these embeddings over non-contextual fastText alternatives \cite{ulvcar2019high}. In another study, Ulcar \& Robnik-Šikonja (2020) constructed trilingual BERT-like models that outperformed large multilingual BERT models on various tasks, showcasing the effectiveness of smaller models tailored for specific languages or language groups \cite{ulvcar2020finest}.

In summary, research has consistently shown that compact BERT models can deliver high performance on language understanding tasks with reduced computational demands. These findings present a compelling case for exploring their application in Turkish language tasks, including sentiment analysis \cite{acikalin2020turkish}, offensive language detection\cite{ozberk2021offensive}, medical text classification\cite{ccelikten2021turkish}, Part-of-Speech tagging \cite{stefan_schweter_2020_3770924}, and Named Entity Recognition \cite{ccarik2022twitter}.

\section{Methodology}
Our study presents the development and evaluation of smaller Turkish BERT models, namely tiny, mini, small, and medium models. Unlike typical approaches that use knowledge distillation to create smaller models, we instead chose to directly train these models with fewer parameters. The goal of this approach was to create more computationally efficient models while maintaining high performance.

In the following sections, we describe the architecture of these models, the extensive dataset used for training, the specific training procedures employed, and the various evaluation tasks used to measure the performance of the models.

\subsection{Model Architecture}

Our study examined the use of BERT models of different sizes. The architecture of each model varied in terms of hidden size, number of attention heads, and number of hidden layers. Table 1 summarizes these architectural differences.

\begin{table}[ht]
\centering
\begin{tabular}{ccccc}
\toprule
Model & Hidden Size & Num. Attention Heads & Num. Hidden Layers & Num. Parameters (millions)\\
\midrule
Tiny & 128 & 2 & 2 & 4.6m \\
Mini & 256 & 4 & 4 & 11.6m \\
Small & 512 & 8 & 4 & 29.6m \\
Medium & 512 & 8 & 8 & 42.2m \\
Base & 768 & 12 & 12 & 110.7m \\
\bottomrule
\end{tabular}
\caption{Architectural differences and parameter counts among the models.}
\label{table:1}
\end{table}

Despite the variations in architecture, all models share some commonalities. They utilize the gelu activation function, which provides a smoother gradient than ReLU. The initializer range is set to 0.02, dictating the range from which the initial model parameters are randomly selected. The vocabulary size is 32000. We used BerTurk's tokenizer. A hidden dropout probability of 0.1 is used to prevent overfitting. Min Vocabulary Size is  refer to the minimum token frequency required for inclusion in the vocabulary. Lastly, a maximum position embedding of 512 is employed, limiting the input sequence length to 512 tokens.

\subsection{Training Data and Process}
The models were trained using a diverse dataset totaling over 75GB in size, which was composed of OSCAR, MC4, the latest Wikipedia data (about 1GB), a collection of clean news data (about 3GB), and a corpus of Turkish and translated novels (about 3GB). While OSCAR and MC4, which are composed of unfiltered crawled data from the internet, contain somewhat contaminated data such as encoding defects and toxic data, the Wikipedia, news, and novel datasets we used are much cleaner. The only filter applied to the datasets was a minimum sentence length of five words. For tokenization, we employed the tokenizer of the BERTurk-uncased model.

We utilized the Huggingface library and TensorFlow for the training process. Each model was trained using the Adam optimizer, with a learning rate of 1e-4 and a batch size of 128.

\subsection{Evaluation Tasks}
To evaluate the performance of our models, we selected common tasks: mask prediction, sentiment analysis, news classification and zero shot classification. These tasks were chosen due to their widespread use in the evaluation of Turkish BERT models. In addition to the task performances, we also report the execution time for each model to provide insight into their computational efficiency.

\section{Experiment}

This experimental section deals with the performance and efficiency of the different Turkish BERT models developed, using the standard base BERT model as a comparative benchmark. The investigations were based on three predominant NLP tasks: mask prediction, sentiment analysis, and news classification. The goal was twofold - to evaluate the performance of our custom models and to compare their efficiency with the base BERT model.

Our suite of models for the experiment included the Tiny, Mini, Small, and Medium Turkish BERT models, complemented by the Base BERT model to establish a performance baseline.

Our study is based on two distinct datasets: the 'News' and 'Sentiment' datasets. Both datasets were subject to extensive preprocessing to tailor them to our research objectives.

The 'News' dataset comprises 979 current news articles sourced from various outlets, spread across six different categories. To ensure the models' performance is unbiased, we only included news articles published exclusively after July 19, 2023, which are not in the training dataset. This is an essential measure in ensuring the models' performance evaluation is based on unseen, current data. For the mask prediction task, a subset of 130 news articles was created through character filtering. In contrast, the entire dataset was utilized for classification tasks.

On the other hand, the 'Sentiment' dataset consists of 8491 examples, which are divided into two sentiment-based classes. After the preprocessing and filtering stages, we utilized only 1000 examples for our study. Each example represents a statement that has been classified based on the sentiment it embodies.

Preprocessing steps for both datasets included converting the text to lower case and normalizing the data. For the mask prediction task, an additional step was introduced: filtering samples by character length. Only samples with a character length ranging from 150 to 512 were retained, ensuring the model had ample context for mask prediction. Notably, this character length filtering was not used in the classification task.

\subsection{Mask Prediction}

The mask prediction task served as the initial phase of the study. This task evaluates a model's ability to decipher context. The process entailed tokenizing the text and replacing five randomly chosen tokens per example with a '[MASK]' placeholder. Subsequently, the models were tasked to predict the masked tokens. This approach yielded five predictions for each masked example, thereby enabling the calculation of top-1 and top-5 accuracies.

The experiment evaluated five distinct models: 'Tiny', 'Mini', 'Small', 'Medium', and 'Base'. These models were selected due to their varied complexity and differing quantities of parameters, ranging from the 'Tiny' model's 4,607,360 parameters to the 'Base' model's 110,650,880 parameters.

As detailed in Table \ref{tab:model_performance}, there was a noticeable trend of incremental improvement in top-1 and top-5 accuracy from the 'Tiny' model to the 'Base' model across both datasets. This pattern underscores the correlation between the increased model complexity, parameter size, and improved model performance. However, this increase in performance necessitates greater computational time.

Specifically, in the News dataset, the 'Base' model required approximately 5.4 times the computational time of the 'Tiny' model. However, it correspondingly showed a significant increase in top-1 accuracy, from 34.15\% with the 'Tiny' model to 79.54\% with the 'Base' model. A similar pattern was observed in the Sentiment dataset, with the 'Base' model necessitating nearly 4.9 times the computational time of the 'Tiny' model, but leading to a substantial improvement in top-1 accuracy from 22.44\% to 51.98\%.

Consistently across all models, the News dataset yielded higher accuracy rates than the Sentiment dataset. This can likely be attributed to the structured and less ambiguous language typically used in news articles, which simplifies prediction tasks for models.

\begin{table}[h]
\centering
\begin{tabular}{|l|l|l|l|l|}
\hline
\textbf{Model} & \textbf{Dataset} & \textbf{Run Time (seconds)} & \textbf{Top-1 Accuracy} & \textbf{Top-5 Accuracy} \\ \hline
Tiny & News & 12.81 & 34.15\% & 47.54\% \\ \hline
Tiny & Sentiment & 69.38 & 22.44\% & 36.44\% \\ \hline
Mini & News & 16.79 & 50.77\% & 66.46\% \\ \hline
Mini & Sentiment & 135.47 & 31.56\% & 49.38\% \\ \hline
Small & News & 26.13 & 61.23\% & 74.46\% \\ \hline
Small & Sentiment & 189.99 & 40.52\% & 60.32\% \\ \hline
Medium & News & 45.44 & 65.69\% & 80.15\% \\ \hline
Medium & Sentiment & 210.89 & 45.10\% & 64.86\% \\ \hline
Base & News & 69.26 & 79.54\% & 91.08\% \\ \hline
Base & Sentiment & 343.09 & 51.98\% & 72.74\% \\ \hline
\end{tabular}
\caption{Performance of different models on the News and Sentiment datasets}
\label{tab:model_performance}
\end{table}

\subsection{Classification Tasks}

In this subsequent phase of the study, the analysis extended into classification tasks with a primary emphasis on sentiment analysis and news classification. The evaluation parameters continued to be analogous to those in the previous phase, maintaining a comparative approach against the base BERT model as a benchmark.

For this phase, each of the five models was employed to convert sentences into vector embeddings. These embeddings were then used as input for a straightforward sequential model composed of three dense layers and a final output layer, tailored to the number of classes in the dataset. This model was trained to execute the classification tasks.

The classification process was repeated ten times for each model. The performance metrics utilized in this stage were the averages of the highest test accuracies achieved across these ten repetitions. Since most processes took less than ten seconds, the times were not included to avoid misleading information due to noise.

The table \ref{tab:classification_performance} presents the performance of each model on both datasets. A consistent pattern, similar to the mask prediction task, is observed: the models' performance on both sentiment analysis and news classification tasks improves with increasing complexity. The 'Base' model consistently demonstrated superior average accuracy across both datasets, although it came with a greater computational burden. In terms of standard deviation, the 'Base' model also displayed lower values, indicating a more reliable and consistent performance across the different runs of the experiment.

This comprehensive exploration of various models across distinct classification tasks emphasizes the balance between efficiency and performance and provides valuable insights into their potential applications.

\begin{table}[H]
\centering
\begin{tabular}{|l|l|l|l|}
\hline
\textbf{Model} & \textbf{Dataset} & \textbf{Mean Accuracy} & \textbf{Std Deviation} \\ \hline
Tiny & Sentiment & 0.8251 & 0.0017 \\ \hline
Mini & Sentiment & 0.8511 & 0.0028 \\ \hline
Small & Sentiment & 0.8882 & 0.0027 \\ \hline
Medium & Sentiment & 0.8973 & 0.0028 \\ \hline
Base & Sentiment & 0.9185 & 0.0015 \\ \hline
Tiny & News & 0.7151 & 0.0362 \\ \hline
Mini & News & 0.7571 & 0.0396 \\ \hline
Small & News & 0.7796 & 0.029 \\ \hline
Medium & News & 0.7944 & 0.0207 \\ \hline
Base & News & 0.8701 & 0.0135 \\ \hline
\end{tabular}
\caption{Performance of different models on the Sentiment Analysis and News Classification tasks}
\label{tab:classification_performance}
\end{table}

\subsection{Zero-shot Classification}

Next phase of our study ventured into the realm of zero-shot classification tasks. This classification technique involves utilizing the trained models to predict classes of data they haven't been explicitly trained on, hence 'zero-shot'. Again the focus was on two primary tasks: sentiment analysis and news classification.

In the sentiment analysis task, we treated the problem as a comparison of sentence similarity: "Bu metnin içerdiği duygu çoğunlukla \{ \}." ("The sentiment expressed in this text is mostly \{ \}."), where "\{ \}" was filled with either 'olumlu' (positive) or 'olumsuz' (negative). We then compared the cosine similarity of sentences in the dataset to this constructed sentence.

The news classification task was structured in a similar way, with the reference sentence: "Bu haberin içeriği çoğunlukla \{ \} ile ilgilidir." ("The content of this news is mostly about \{ \}."), with "\{ \}" being filled with one of six news categories: 'dunya' (world), 'ekonomi' (economy), 'kultur-sanat' (art-culture), 'magazin' (entertainment), 'politika' (politics), and 'spor' (sports). The dataset sentences were then compared to these constructed sentences using cosine similarity.

For both tasks, the models' sentence vectors were used to measure their cosine similarity to the vectors derived from the reference sentences. The class label of the reference sentence with the highest similarity was then assigned to the dataset sentence.

The prompts utilized in our sentiment analysis and news classification tasks have been adapted from a previous study \cite{ccelik2023unified}.

\begin{table}[h]
\centering
\begin{tabular}{|c|c|c|}
\hline
\textbf{Model} & \textbf{Dataset} & \textbf{Zero-shot Accuracy} \\ \hline
Tiny & Sentiment & 55.41 \% \\ \hline
Mini & Sentiment & 60.57 \% \\ \hline
Small & Sentiment & 68.25 \% \\ \hline
Medium & Sentiment & 67.68 \% \\ \hline
Base & Sentiment & 82.16 \% \\ \hline
Tiny & News & 27.89 \% \\ \hline
Mini & News & 26.86 \% \\ \hline
Small & News & 25.13 \% \\ \hline
Medium & News & 28.70 \% \\ \hline
Base & News & 32.69 \% \\ \hline
\end{tabular}
\caption{Performance of different models on the Zero-shot Classification tasks}
\label{tab:zeroshot_classification_performance}
\end{table}

As demonstrated in Table \ref{tab:zeroshot_classification_performance}, there is a conspicuous upward trend in the performance of the zero-shot classification tasks as we move from simpler to more complex models, with the 'Base' model leading the pack. Interestingly, the performance across the smaller models ('Tiny', 'Mini', 'Small', and 'Medium') displayed relatively little variance, underlining the potential of these less complex models in real-world applications where computational resources might be limited.

In terms of specific tasks, sentiment analysis consistently outperformed news classification. This is likely due to the fewer classes involved in the sentiment analysis task, making it inherently less complex for models to handle in a zero-shot context.

This final phase underscores the fact that while model complexity can bolster performance, the returns diminish as we move from simple to complex models, with smaller models already performing admirably in a challenging zero-shot scenario. This is an important finding for practical applications, as it supports the argument for the adoption of simpler models where resources are at a premium.

\subsection{Sentence Vectorization Time Analysis}

The research expanded to include a test of computational efficiency, measured by the time each model took to vectorize sentences. For generating sentence embeddings, we employed the methodology outlined in the SentenceTransformers \cite{reimers-2019-sentence-bert}. The vectorization of sentences, a fundamental step for most applications, was examined using a distinct dataset consisting of 57,940 news samples \cite{TurkishNewsDatasetKaggle}. All the measurements were conducted on an NVIDIA RTX 3060 GPU with a batch size of 32.

\begin{figure}[h]
\centering
\includegraphics[width=\textwidth]{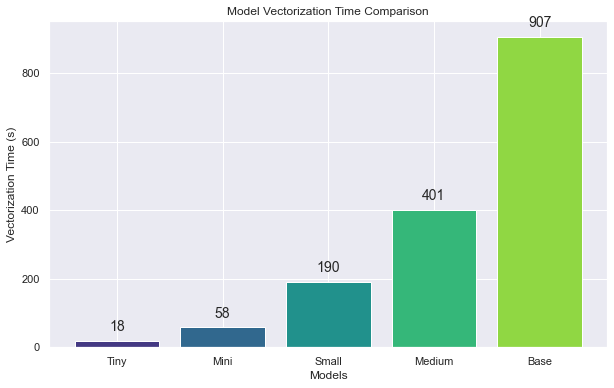}
\caption{Comparison of Vectorization Time for Different Models}
\label{fig:vectorization_time}
\end{figure}

The times recorded for the 'Tiny', 'Mini', 'Small', 'Medium', and 'Base' models are documented. This data is represented visually in Figure \ref{fig:vectorization_time}.

Observations reveal an exponential increase in the time consumed as the model complexity increases. Specifically, the 'Tiny' model vectorized sentences about 50 times faster than the 'Base' model. This showcases the substantial trade-off between computational efficiency and performance in more complex models. These findings, therefore, underscore the importance of selecting the model that most aptly balances the efficiency-performance trade-off for the specific requirements of the application in question.

\section{Conclusion}
In this research, we took a deep dive into the less studied area of creating and analyzing more compact Turkish BERT models. We designed a series of models, varying from tiny to medium in size, and tested them on diverse tasks, including mask prediction, sentiment analysis, and news categorization.

Our studies underscored a valuable balance offered by the smaller models, presenting an efficient compromise between computational resources and performance, despite the superior results exhibited by the larger Base model. To illustrate, the Base model, despite achieving the best results across all tasks, also required significantly more computational resources and time.

These findings carry great significance for those working in environments where there is a scarcity of computational resources, yet a desire to leverage the advantages of advanced NLP techniques. Smaller models could offer a feasible and efficient solution in such scenarios. Our results also underscore the potential of utilizing smaller models for tasks demanding quicker processing and lower memory usage.

We anticipate our research will spur additional exploration into the development and optimization of smaller language models, particularly for languages that have been less represented in NLP research. As a part of our future work, we are planning to experiment with knowledge distillation methods to further strike a balance between the performance and efficiency of our models.

With our endeavor to reduce the size of BERT models, we are striving to make this cutting-edge language model more accessible for tasks demanding quicker processing and lower memory usage, with negligible compromise on performance. Our findings offer precious insights for researchers and professionals in the NLP field, especially within the Turkish language context, setting the groundwork for the development of smaller and more efficient language models in the future.

In conclusion, this study signifies an important leap towards extending the benefits of modern NLP technology to more languages and a broader range of applications.

\section*{Acknowledgments}
We thank the [Google TPU Research Cloud](https://sites.research.google/trc/about/) program for providing part of the computation resources.

\bibliographystyle{unsrt}  
\bibliography{references}

\end{document}